# Automatic Rule Learning for Autonomous Driving Using Semantic Memory*

Dmitriy Korchev, Aruna Jammalamadaka, and Rajan Bhattacharyya

*Abstract*— This paper presents a novel approach for automatic rule learning applicable to an autonomous driving system using real driving data. We represent the actions of other agents (provided by sensors) in the scene via temporal sequences called "episodes". The proposed method adaptively creates new rules automatically by extracting and segmenting valuable information about other agents and their interactions. During the training phase, the system automatically segments driving episodes and extracts rules from real driving data. These rules, which take the form of a "spatiotemporal grammar" or "episodic memory" are stored in a "semantic memory" module for later use. During the testing phase, the system segments constantly changing situations, finds the corresponding parse tree for the current state of the self-car and other agents, and applies the rules stored in semantic memory to stop, yield, continue driving, etc. The method also allows for continues online training during agent driving. Unlike traditional deep driving and machine learning methods that require significant amount of training data to achieve desired quality, the proposed method demonstrates good results with just a few training examples. The system requires proper selection of training examples to avoid learning incorrect driving rules like missing a stop sign, running a red light, etc. The proposed method can provide a set of options available to the path planning module for making control decisions to resolve complex driving situations.

INTRODUCTION

This paper proposes an approach for automatic (unsupervised) rule learning applicable to autonomous driving. The rules are segmented and extracted from the data without prior information or labeling by a user. Current approaches use rule based systems [1] that require significant effort in describing these rules in the system. This approach is different from traditional rule based ones because it removes the tedious process of manual creation of these rules that the system needs to follow during driving in different situations, conditions, and interactions with other agents in the scene.

Unlike traditional machine learning based methods including deep learning [2] that require significant amount of training data to achieve desired quality, our method demonstrates good results with just a few training examples. The advantage of our approach is demonstrated in the paper on a few examples showing that the driving rules can be reliably learned from a very small data set covering desired scenarios.

The rules learned by the proposed system have representation that is easily understandable by humans unlike other machine learning based methods. Another advantage of the proposed system is that these rules can be easily modified, altered, added or removed from the system by a human with minimal re-training. In addition to these advantages the proposed approach can correctly handle unseen situations. All the claimed properties of the proposed approach are demonstrated on a few examples below.

The main idea of the proposed approach is the construction of the semantic memory located in the cognitive module and it's based on the idea proposed in [3] that handles learning the rules for the spatial and temporal events of the autonomous driving system. This cognitive module will be considered in details below as a part of the high-level architecture for autonomous driving. The paper is organized as the following: Section II provides overall system architecture; details on cognitive module are considered in Section III; examples of two driving scenarios demonstrating the functionality of the cognitive module are presented in Section IV.

SYSTEM ARCHITECTURE

The high-level block-diagram of the autonomous driving system is shown in Figure 1 and it is presented here to demonstrate the place of the Cognitive Module shown in yellow on it.

**Cognitive Module**. This module is trained to provide data for Motion Control Module during autonomous driving. It uses the current state and future predictions of the agents in the scene, map data, self-motion planner data, and motion control data. This module contains sub-modules called "semantic" and "episodic" memories, the details for which are provided below.

*Research supported by GM Corporation.

All authors are with HRL Laboratories, LLC, Malibu, CA 90265 USA (Corresponding author is D. Korchev, phone: 310-317-5525; e-mail: dvkorchev@ hrl.com).

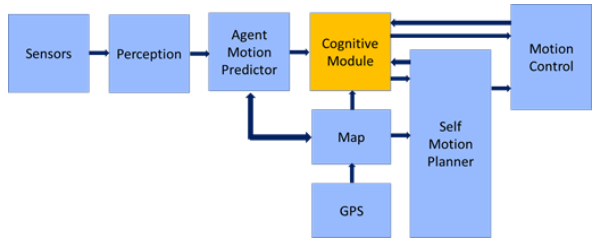

Figure 1 Block diagram of the proposed system for autonomous driving

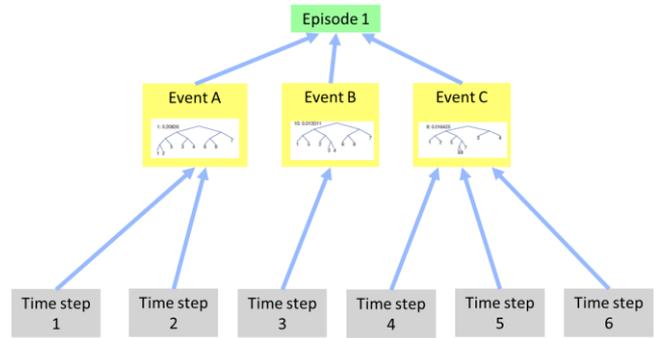

Figure 2 Hierarchy of objects inside a single episode

## COGNITIVE MODULE

**Cognitive Model**. Before going into the details of the Cognitive Module and its functionality, we need to consider the relationships between the driving scenarios that can be broken into episodes, events, and time steps. In this context the driving process is considered as a sequence of episodes, where each episode represents a specific realization of the driving situation. For example, crossing a 4-way stop intersection is a driving situation that can be navigated in many ways depending on presence of other cars, pedestrians, and other actors in the scene. Each realization of this driving situation is called an episode. Therefore, an episode consists of a number of events where each event represents a specific action of the self-car and other agents. For example: Event A – self-car is moving straight, Event B – self-car stops, Event C – self car yields to other agents, and so on. The hierarchy of the components comprising a single episode is shown in Fig. 2. Every event has a time duration in time steps sampled at a specified interval T. Number of time steps in the event depends on speed of the self-car as well as speed and number of other agents, duration of traffic lights, traffic, and other factors. The number of events in episodes correspond to the same traffic situation can also vary for the same reasons. Depending on a particular behavior and number of other agents, the same intersection can be crossed with a different number of events and time steps, comprising different episodes.

The grouping of time steps into events is based on the structure of the parse trees of the time steps (the details of creation of these parse trees are described below). The time steps with identical parse trees represent the same event, this is shown in Figure 2, and a parsing tree for each event is shown inside the yellow boxes.

An example of a group of episodes for a driving situation is shown in Fig. 3. Each episode is broken into events and each event consists of multiple time steps. This segmentation of events inside the episode is performed automatically by the semantic memory of the proposed system. During the training phase the semantic memory will learn two essential models: (1) the transition matrix (3D tensor) for generation of parse trees for the time steps; (2) the temporal grammar that captures temporal relationships between events in the episode. During the testing phase, the semantic memory will create a parse tree for each time step to associate it with the corresponding event, build a sequence of events, and apply temporal Probabilistic Context-Free Grammar (PCFG) to predict the rest of the sequence of the events in the episode.

In other words, model (1) is directed to the spatial relationships/interactions, model (2) is the temporal sequence of events.

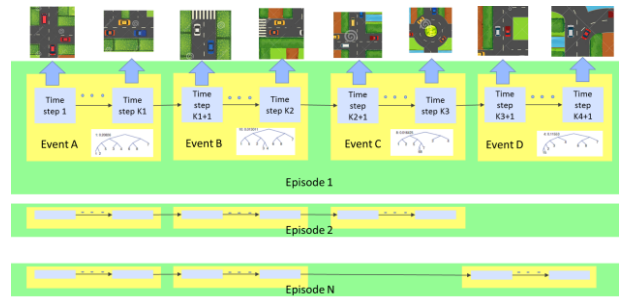

Figure 3 Relationships between driving episodes, events and time steps for a single driving situation

The parse trees in model (1) defined by equation (1) represent events where each events is a specific action of the self-car like: driving forward, waiting on a stop sign, yielding to incoming or cross traffic, etc.

**Spatial Grammar**. The segmentation of events is performed in the cognitive module that contains the both the spatial and temporal parts of the semantic memory. Fig. 4 shows the block diagram of phase 1 training of semantic memory (spatial part of the semantic memory). All training scenarios are broken into time steps representing a snapshot or state of the agents in the scene. Each time step is represented by information describing spatial locations and states of the agents in the scene relative to the self-car and other agents (stop sign, traffic lights/signs etc.) depending on the type of intersection or road. Several detailed examples of this kind of representation will be provided below. The spatial information of the agents in each time step is converted into an "association matrix" [3] which can be considered as a compact representation of that time step and this matrix is used as an input of the training/testing algorithm. Association matrices obtained from each time step are then fed into a perceptual Expectation Maximization (EM) algorithm [3], which is a modified version of the inside-outside algorithm that is usually used for training PCFG.

This training step produces a transition matrix (3D tensor) representing the extracted and optimized probabilistic rules of the spatial interaction between the agents and self-car in the scene. The transitions rules $p(i \rightarrow j, k)$ are modified [3] according the following equation

$$p(i \rightarrow j, k) = a(i,j,k)M(P,Q), \quad (1)$$

where $a(i,j,k)$ represents elements of the semantic transition matrix, $M(P,Q)$ represents set-wise association specific for the current episode, $i, j, k$ – indices of the nodes in the structure representing the grammar.

The transition matrix can be used to generate the most probable parse trees for the time steps used for training as well as for finding the optimal parse trees during testing. The unseen time steps will produce the parse trees that show how self-car interacts with other agents during this time step.

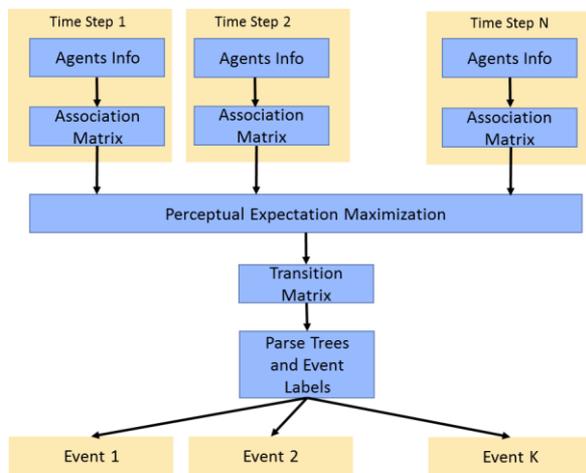

Figure 4 Semantic memory, training phase 1, and single scenario event segmentation

In summary, after the creation of the transition matrix, semantic memory is able to generate the most probable parse tree for each time step in the training set, based on the rules and probabilities in the grammar. Generation of the most probable parse trees for the test set containing new combinations of agent interactions in the scene (previously not seen during the training phase) will be described in detail below in testing phase of the semantic memory. In some situations several parse trees are possible with very close probability; these situation should to be resolved with the path planning module that will decide what action is more suitable for the self-agent.

**Temporal Grammar**. After training phase 1 (spatial rules) the cognitive module goes to the next phase of training where it learns the temporal relationships between the events in the episodes provided in the training data set This process results in the temporal sematic memory. Since many time steps have the same optimal parse tree, each episode can be represented as a much shorter sequence of so called events. For example, the episode has parse trees corresponding the following sequence of time steps: {AAAABBBBCCCC}. After removing duplicates, the episode can be represented as shorter sequence of events: {ABC}, similar to run length encoding compression. The number of repetitions of each time step depends on a specific realization of the episode, speed of agents, traffic situation, etc. But all these variations will have the same sequence of events: {ABC}, where A, B, and C are the events in this episode with the corresponding parse trees. In other words, the event consists of time steps in which the self-car does not change its action and interaction with other agents. For example, event A represents self-car driving forward, event B represents different situation when self-car stops, event C – self-car yields to other agents in the scene, etc.

The block diagram of phase 2 training of the semantic memory is shown in Fig. 5. Each episode in the training data is represented as a sequence of events. In other words, each episode is represented as a string of words (events). All these strings form a corpus of strings that can be stored in episodic memory for future retrieval or can be used for learning a temporal PCFG, which is a generative model that can be used for predictions of events in the future episodes. The algorithm for learning a PCFG is presented in [4]; the approach in [3] is also suitable for this purpose. In addition to the learned PCFG, the cognitive module can store all unique sequences in episodic memory and use those for predictions during the testing phase as well.

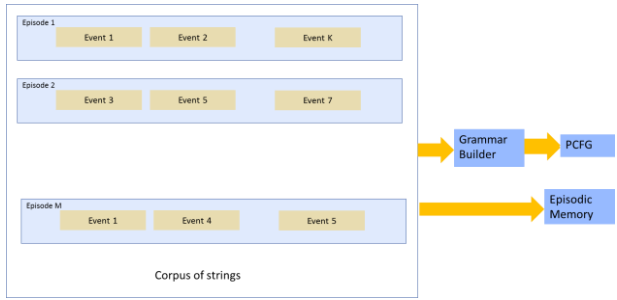

**Figure 5** Semantic memory, training phase 2, and learning temporal relationships between events in episodes in form of temporal Probabilistic Context Free Grammar (PCFG)

Block diagram of semantic memory during testing phase is shown in Fig. 6. The module processes data in time steps similar to the training phase. Each time step contains spatial information and state for each agent in the scene. This information is converted into the association matrix which is processed with the transition matrix (obtained during training phase1) to find the optimal parse tree for the current time step. This optimal parsing can be found with an algorithm presented in [5]. The optimal parse tree for the current time step is compared to the set of most probable parse trees learned during the training to find the matching one and assign an event label to the current time step. Sequential duplicate time step labels are discarded. This process will form a temporal string of labels corresponding to the evolving episode in the driving scenario. The prediction of the rest of the episode can also be performed using episodic memory which holds all previously seen episodes. In this case, predictions are made by finding the matching or closest sub-sequence of the events representing the current unfinished episode against all episodes stored in the memory.

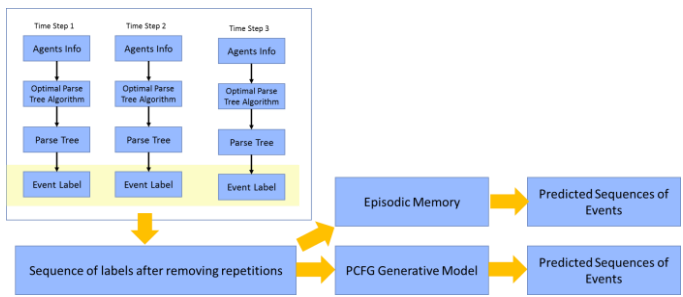

**Figure 6** Semantic memory, testing phase. The temporal predictions can be generated by either PCFG and/or episodic memory

The matched episodes are used to predict the set of events that will follow to complete the current episode. If a PCFG is used the prefix string containing event labels corresponding to the current partial episode is used to predict the rest of the sequence using rules and probabilities learned during training PCFG.

Prediction of future events in the evolving episode is performed at every time step. The predictions are provided to the motion planning module to execute optimal control actions. For example, in the presence of a stop sign on a street, all episodes will contain an event corresponding to the self-car stopping, regardless whether or not the other traffic is present.

FUNCTIONAL EXAMPLES OF COGNITIVE MODULE

**Left Turn Scenario on a 2 way street**. In order to further elucidate our method, we consider a left turn situation where a self-car drives straight and makes a left turn to a 2-way street. The self-car may stop to yield to incoming traffic. This scenario is shown in Fig. 7. There are 6 elements of self-car context that represent this scenario, they are shown on the left side of the Fig. 7

Self-car context for the driving scenario 1:
1 – self-car drives forward
2 – self-car stops to yield the incoming traffic
3 – self-car turns left
4 – self-car turns right
5 – car on right moving in the same direction
6 – car on left moving in opposite direction representing incoming traffic

In general case self-car context can include not only the positions of the agents (vehicles) but also goals, and actions of other vehicles.

The right side of the Fig. 7 shows trajectories of each car, the red dot is a stopping point for the self-car to yield the incoming traffic.

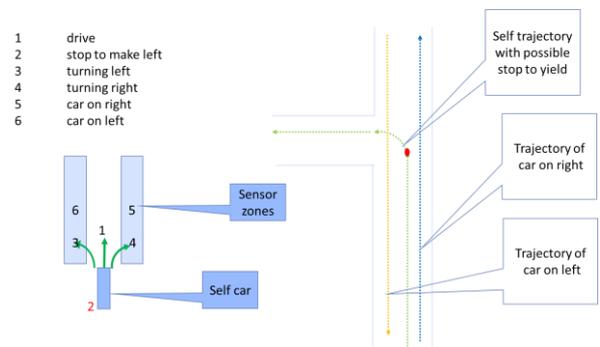

**Figure 7** Semantic memory, left turn, scenario 1. Green part of self-trajectory indicates that car is moving, the red dot is a stop point to yield to incoming traffic

The first group of time steps 13-17 of the left turn is shown in Fig. 8. Since the first phase of training requires only time steps from multiple episodes, in this context the group of time steps does not represent all time steps (a complete episode) from the beginning to the end of the left turn. The first step of training does not require any order of the time steps because it finds the relationships between objects inside these time steps. The location of a single incoming car at each time step is shown on the right as yellow rectangles with the corresponding time step number inside. The left part of the

Fig. 8 shows information that is used for creation of the association matrices for each time step. The first row defines state of the self-car (in this case stopping to make a left, turning left, with a car on left) as defined by the numbers in Fig. 7, the second one is the vertical distance to the self-car, the third row is the horizontal distance to the self-car, for simplicity all horizontal distances are zeros.

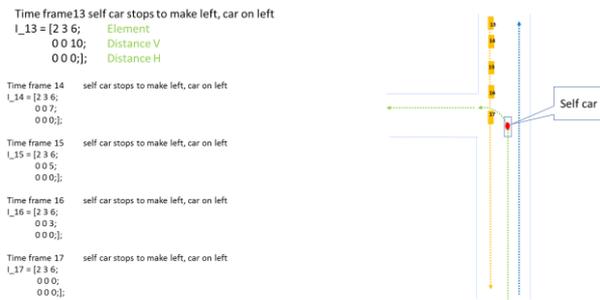

**Figure 8 Left turn, scenario 1, time steps 13-17**

Fig. 9 shows time steps 18-22 of the left turn scenario. In this time steps the self-car is moving forward to the left turn point. Car on left represents incoming traffic, car on right represents traffic that moves in the same direction but one lane to the right.

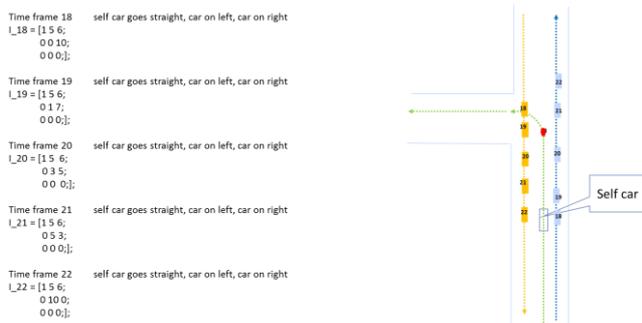

**Figure 9 Left turn, scenario 1, time steps 18-22**

Fig. 10 shows time steps 23-27 for the left turn scenario. In these time steps the self-car stops to yield to incoming traffic represented by self-car context 6. The self-car context 5 on the right represents traffic moving in the right lane in the same direction.

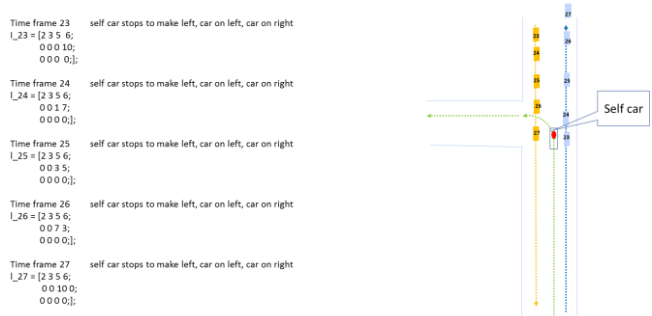

**Figure 10 Left turn, scenario 1, time steps 23-27**

Association matrices are fed into an EM optimization to produce a grammar, and corresponding parse trees per episode, as described above. The resulting most probable parse trees generated by the semantic memory during training of phase 1 for the driving situation 1 are shown in Fig. 11. These trees represent all previously described time steps. We can see that many of these trees are duplicates.

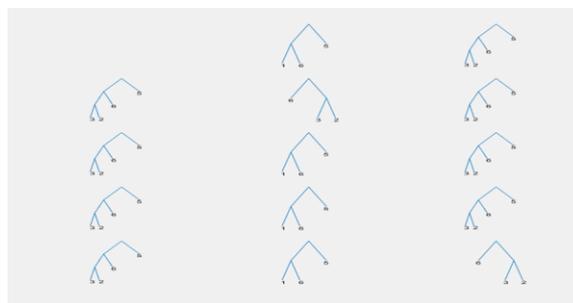

**Figure 11 Left turn, most probable parse trees for scenario 1**

Fig. 12 shows the grouping of the parse trees into significantly smaller set representing possible events in the left turn scenario 1. We can see that scenario 1 can be described by three types of events. These three types of event will define the temporal grammar for the specified type of left turn.

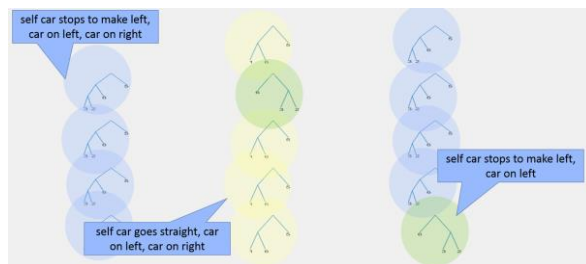

**Figure 12 Left turn, parse trees for scenario 1, grouping of the trees into events**

**Left Turn Scenario on a T-Intersection with a stop sign**. Let's consider more complex left turn scenario 2 with a stop sign at a T-type intersection, shown in Fig. 13. This case will have cross traffic from left and right, and traffic in the same direction in the lane to the right. All agents of this scenario are shown on the left part of Fig. 13. Compare to the first scenario we have three new elements: traffic crossing from the right, traffic crossing from the left, and stop sign.

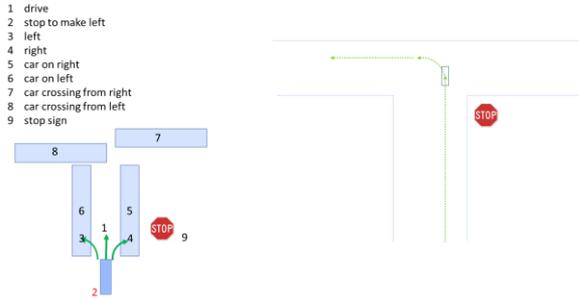

Figure 13 Semantic memory, left turn, scenario 2 with a stop sign. Green line represents the trajectory of self-car.

Time steps 18-22 of the scenario 2 are identical to that of scenario 1 and they are shown in Figure 14. The self-car is moving straight, incoming traffic is represented by car 6 and same direction traffic is represented by self-car context 5 moving in the right lane.

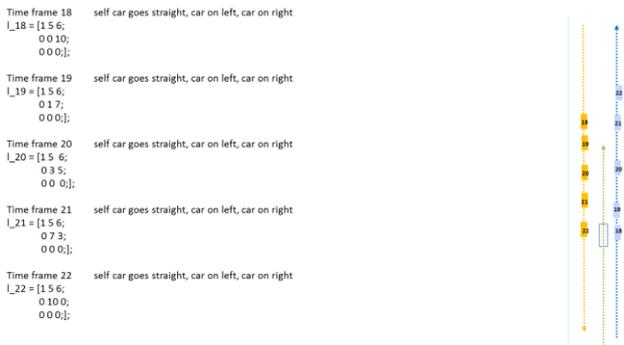

Figure 14 Left turn, scenario 2, time steps 18-22

Time steps 31-34 for the scenario 2 are shown in Fig. 15. In this time steps the self-car stops at the stop sign and waits for the traffic crossing from left. Only horizontal distance between self-car and crossing car is used for creation of the association matrices, for the simplicity, the vertical distances are zeros.

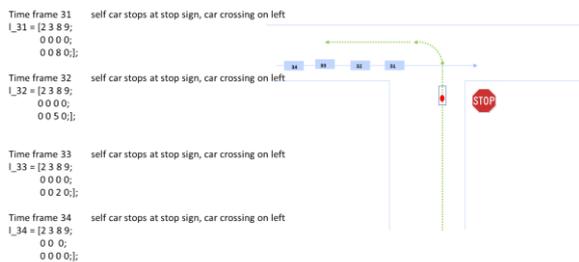

Figure 15 Left turn, scenario 2, time steps 31-34

Time steps 35-39 for scenario 2 are shown in Fig. 16. These time steps show that the self-car stops at the stop sign and waits for the traffic crossing from right. Only horizontal distances between self-car and crossing car are used for creation of the association matrices, for simplicity, the vertical distances are zeros.

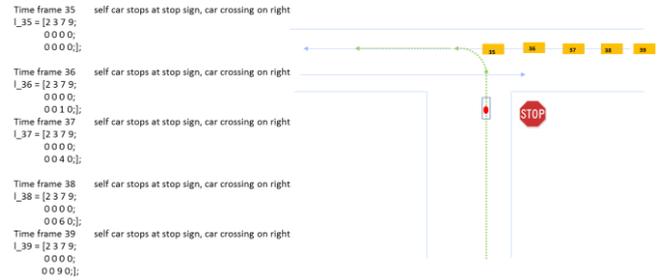

Figure 16 Left turn, scenario 2, time steps 35-39

Time steps 40-44 for left turn scenario 2 are shown in Fig. 17. In these time steps the self-car stops at the stop sign and waits for the traffic crossing from both left and right. Only horizontal distances between self-car and crossing cars are used for the association matrices, all vertical distances can be considered close to zeros and they do not change over time steps.

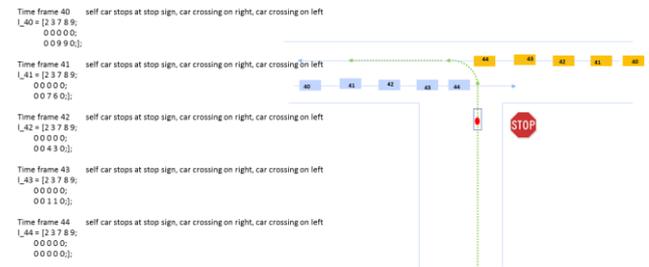

Figure 17 Left turn, scenario 2, time steps 35-39

Time step 45 for left turn scenario 2 is shown in Figure 18. In this time step the self-car makes the left turn after a stop, and there is a car on the right that affects the behavior of the self-car depending on each car's arrival time to the stop sign.

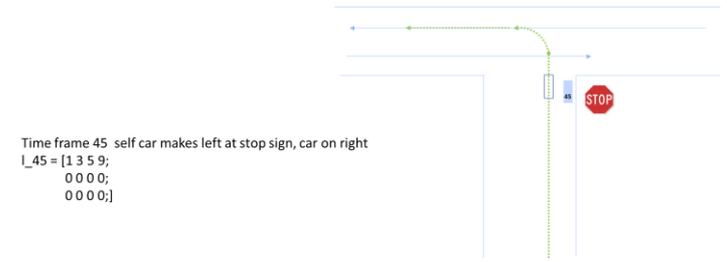

Figure 18 Left turn, scenario 2, time step 45

Time step 47 of the left turn scenario 2 is shown in Figure 19. In this time step the self-car stops at the stop sign. No other car is present at the intersection in this time step.

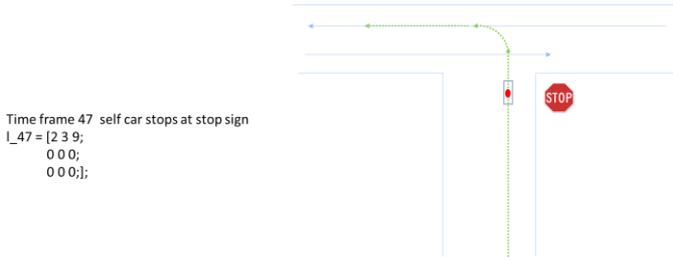

**Figure 19 Left turn, scenario 2, time step 47**

Fig. 20 shows most probable parse trees for the left turn scenario 2. These trees were generated during phase 1 training of the semantic memory. Like the previous scenario, we can see that many of these trees are also duplicates.

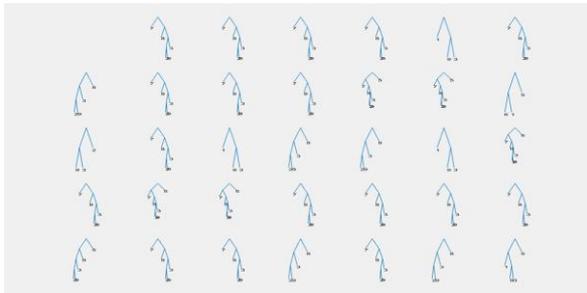

**Figure 20 scenario 2, parse trees**

The colors in Fig. 21 show the grouping of the parse trees into significantly smaller subset of 7 trees, representing the events in the episodes for the left turn scenario 2. We can see that the semantic memory generated two new events that were not in the training set before: self-car stops, car on right, cars crossing from left and right; self-car stops, car on right, car crossing from right. These new events do not contradict the driving rules learned during training: the self-car stops and waits for crossing traffic. This combination of agents with a car on right was not a part of the time steps in the training set. The semantic memory assigned correct interaction to the self-car for this new unseen event.

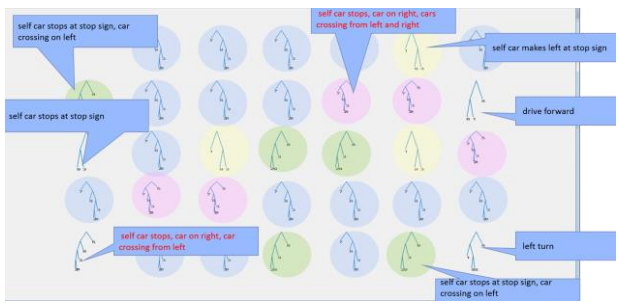

**Figure 21 scenario 2, grouping of the parse trees into events**

The new generated events are shown in Figure 22 and Figure 23.

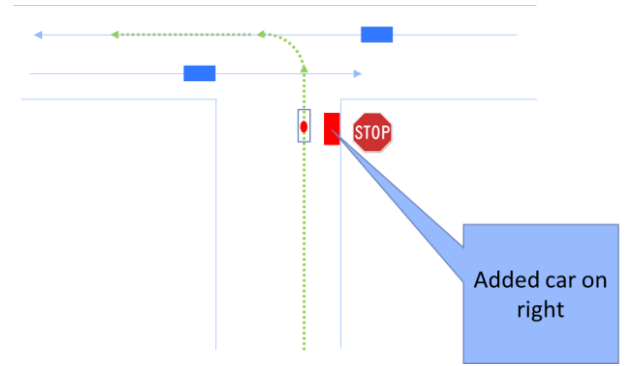

**Figure 22 New event generated from transition matrix.**

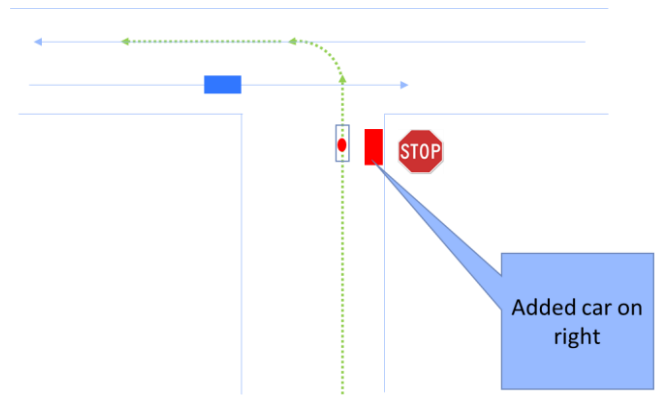

**Figure 23 Another new event generated from transition matrix**

The approach presented above can be used to handle different types of intersections and traffic situations like a 4-way stop intersection and other complex scenarios.

Handling multiple objects. The examples above show how the semantic memory handles only the "representative" agents in the episodes which require the self-car's immediate attention. In real driving conditions there are many possibilities that cars will be moving in groups representing incoming, crossing, and other types of traffic in the scene. In this case, the semantic analysis is applied to multiple combinations of the agents with the constraint that only one object from a particular group is used. For example, if we have three cars in incoming traffic for the scenario 1, we analyze each car from this group separately. The groups can be defined by the layout of driving lanes in the scenario.

## CONCLUSIONS

The examples above demonstrate how the semantic memory learns rules based on the locations of the other agents in the scene. The agent motion prediction module can generate prediction of each agent for K number of time steps forward. These predicted locations of the agents can be used for processing time steps to train phase 1 of the semantic memory and during the testing phase in the same way as it was described above. The rules learned with the cognitive module can provide valuable information on the set of actions available to the path planning part for the final resolution of which action should be taken by the autonomous agent.